\documentclass[runningheads]{llncs}
\usepackage[T1]{fontenc}

\usepackage{amsmath} 
\usepackage{graphicx}
\usepackage{subfigure}
\usepackage{wrapfig}
\usepackage{caption}
\usepackage{enumitem}
\usepackage{color,marvosym}
\usepackage{graphicx,verbatim}
\begin{document}
\title{Domain-Adaptive Diagnosis of Lewy Body Disease with Transferability Aware Transformer}
%
\author{Xiaowei Yu\inst{1} \and
Jing Zhang\inst{1} \and
 Tong Chen\inst{1} \and Yan Zhuang\inst{1} \and Minheng Chen\inst{1} \and Chao Cao\inst{1}  \and Yanjun Lyu\inst{1}  \and Lu Zhang\inst{2} \and Li Su\inst{3,4} \and Tianming Liu\inst{5} \and Dajiang Zhu\inst{1}\textsuperscript{(\Letter)} }
\authorrunning{X. Yu et al.}
\institute{Department of Computer Science and Engineering, University of Texas at Arlington, USA  \\
\email{\{dajiang.zhu@uta.edu\}}\\
\and
Department of Computer Science, Indiana University Indianapolis, USA \and
Sheffield Institute for Translational Neuroscience, University of Sheffield, UK \and
Department of Psychiatry, University of Cambridge, UK \and
School of Computing, University of Georgia, USA }

\maketitle              
\begin{abstract}
Lewy Body Disease (LBD) is a common yet understudied form of dementia that imposes a significant burden on public health. It shares clinical similarities with Alzheimer’s disease (AD), as both progress through stages of normal cognition, mild cognitive impairment, and dementia. A major obstacle in LBD diagnosis is data scarcity, which limits the effectiveness of deep learning. In contrast, AD datasets are more abundant, offering potential for knowledge transfer. However, LBD and AD data are typically collected from different sites using different machines and protocols, resulting in a distinct domain shift. To effectively leverage AD data while mitigating domain shift, we propose a Transferability Aware Transformer (TAT) that adapts knowledge from AD to enhance LBD diagnosis. Our method utilizes structural connectivity (SC) derived from structural MRI as training data. Built on the attention mechanism, TAT adaptively assigns greater weights to disease-transferable features while suppressing domain-specific ones, thereby reducing domain shift and improving diagnostic accuracy with limited LBD data. The experimental results demonstrate the effectiveness of TAT. To the best of our knowledge, this is the first study to explore domain adaptation from AD to LBD under conditions of data scarcity and domain shift, providing a promising framework for domain-adaptive diagnosis of rare diseases.


\end{abstract}
\section{Introduction}
Lewy Body Disease (LBD) is a progressive neurodegenerative disorder and one of the most common causes of dementia~\cite{Walker_2015}, yet it remains underexplored compared to Alzheimer’s disease (AD). LBD shares clinical similarities with AD, as both diseases exhibit cognitive decline progressing through cognitive normal (CN), mild cognitive impairment (MCI), and dementia stages. For both LBD and AD, once established, they cannot be cured or prevented~\cite{Lyu2023Classification}\cite{Capouch2018}. Therefore, improving diagnostic accuracy is crucial for early intervention to slow disease progression~\cite{Lyu2024Mild}\cite{Zhang2025Classification}\cite{Zhang2025Brain}. Despite this importance, accurate diagnosis of LBD remains a challenge due to the scarcity of LBD data, limiting the effectiveness of deep learning models.

In contrast, Alzheimer’s disease has been extensively studied, with large-scale datasets publicly available~\cite{Yu2021Free}\cite{Zhang2025BrainNet}\cite{Lyu2024Gp}\cite{Lyu2021Classification}\cite{Chen2025Core}. This discrepancy presents an opportunity for knowledge transfer~\cite{Fang2025Knowledge}, where insights gained from AD data can enhance LBD diagnosis. However, a key obstacle in this process is the domain shift, where LBD and AD datasets are usually collected from different institutions, using varying imaging machines and acquisition protocols, leading to distinct differences in data distribution. Conventional deep learning models struggle with such discrepancies, limiting their generalizability across diseases.

To leverage knowledge from a larger AD dataset for effective model training while mitigating the domain shift, we propose the Transferability Aware Transformer (TAT), a novel vision transformer (ViT)-based approach designed to adapt knowledge from AD to improve LBD diagnosis. TAT utilizes structural connectivity (SC) derived from structural MRI as input and selectively emphasizes transferable features while suppressing domain-specific ones. At the core of TAT is the Transferability Aware Self-Attention (TAS) mechanism, specifically designed for ViT to amplify disease-transferable features between LBD and AD while suppressing less transferable ones. TAS employs a local discriminator to assess the transferability of patches by distinguishing whether they originate from LBD or AD datasets, generating a transferability matrix. This adaptively learned matrix is then integrated into the self-attention mechanism, enabling the model to focus effectively on transferable features. Through this process, TAS assigns greater weights to more transferable attention operations while reducing the influence of less transferable features. To further mitigate domain shift, we introduce a global discriminator, which takes the classification token as input to distinguish samples from LBD and AD datasets. By aligning cross-domain representations, TAT leverages the larger AD dataset to enhance LBD diagnosis. Experimental results demonstrate that TAT outperforms state-of-the-art domain adaptation methods, providing a promising framework for LBD diagnosis.

\section{Method}
We model the problem within the domain adaptation framework~\cite{Yu2025Feature}\cite{Yu2023Robust}. Let $D_{s} = \left \{ \left ( x_{i}^{s},y_{i}^{s}  \right )  \right \}_{i=1}^{n_{s} }$ represent samples in the labeled AD dataset, where $x_{i}^{s}$ denotes the structural connectivity (SC), $y_{i}^{s} $ denotes the corresponding labels, and $n_{s}$ is the number of samples. Similarly, let $D_{t} = \left \{ \left ( x_{j}^{t}  \right ) \right \}_{j=1}^{n_{t} }$ represent samples in the LBD dataset, consisting of $n_{t}$ SCs but no labels. Since the number of available LBD samples is limited, it satisfies $n_{t} \ll  n_{s}$. During training, both samples and corresponding labels from the AD dataset are employed, while only the unlabeled samples from the LBD dataset are incorporated.

\begin{figure*}[t]
\centering
\includegraphics[width=\textwidth]{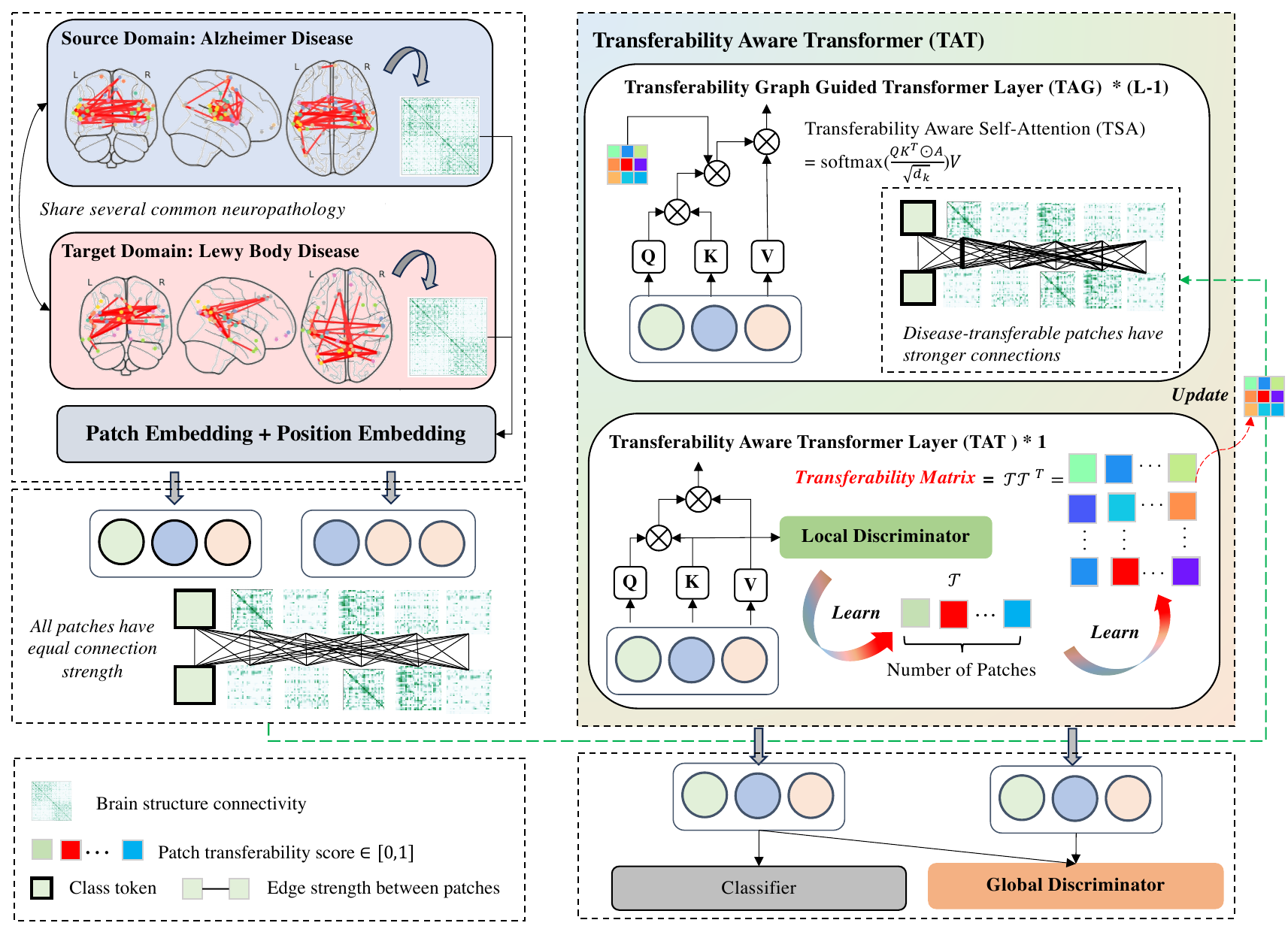}
\caption{Illustration of proposed framework. The TAT layer employs a local discriminator to dynamically assess the transferability score of each SC patch and generate a corresponding transferability matrix. This matrix is then integrated into other Transformer layers to adjust the self-attention strength and thus enhance cross-domain knowledge sharing. The global discriminator utilizes the classification token to align the global features.} \label{method}
\end{figure*}

\subsection{Overview of TAT}
Figure~\ref{method} illustrates the overall framework
of our proposed TAT. TAT is built upon a Vision Transformer (ViT) and comprises a global discriminator that evaluates the entire SCs using class tokens, a local discriminator that assesses the transferability of individual patches, and a classification head. The ViT backbone is enhanced with the transferability-aware attention, which includes both the Transferability Graph-Guided Transformer Layer and the Transferability-Aware Transformer Layer. Following the patch division in ViT~\cite{Yu2024CorePeriphery}\cite{Yu2024CPCLIP}, each SC is divided into non-overlapping patches, which are linearly projected into the latent space and concatenated with positional information. A class token is prepended to the patch tokens, and all tokens are subsequently processed by the backbone. From a graph perspective, we consider the patches as nodes and the attention between patches as edges. This allows us to manipulate the strength of self-attention by adaptively learning a transferability-aware graph (TAG).

\subsection{Transferability Aware Transformer Layer}

The patch tokens correspond to partial of the SC and capture features as fine-grained brain representations. In this work, we define the transferability score on a patch to assess its transferability. A patch with a higher transferability score is more likely to correspond to the highly transferable features across the diseases. 

To obtain the transferability score of patch tokens, we adopt a patch-level local discriminator $D_{l} $ to evaluate the local features with a loss function:
\begin{equation}\label{PatchLevelloss}
L_{pat} (\tilde{z}, y^{d})= -\frac{1}{nP} \sum_{\tilde{z}_{i} \in D} \sum_{p=1}^{P} L_{CE} \left ( D_{l} \left (  \tilde{z}_{ip}  \right )   , y_{ip}^{d}  \right ) 
\end{equation}
where $P$ is the number of patches, $D= \left \{ D_{s}\right \}\cup \left \{ D_{t}\right \} $, $D_{l}$ is the local discriminator, $n =  {\textstyle \sum_{s=1}^{S}} n_{s} + {\textstyle \sum_{t=1}^{T}}n_{t}$, is the total number of SC in $D$, $y_{ip}^{d}$ denotes the macro label of the $p$th token of the $i$th SC, i.e., $y_{ip}^{d} = 1$ means labeled SCs in AD dataset, $y_{ip}^{d} = 0$ the unlabeled LBD dataset. The distributions of features are aligned when the local discriminator finds it difficult to differentiate the macro labels of the patch tokens. $D_{l}\left ( \tilde{z}_{ip} \right ) $ gives the probability of the patch belonging to the labeled samples in the AD dataset. During the adaptation process, $D_{l}$ tries to discriminate the patches correctly, assigning 1 to patches from the labeled data and 0 to those from the unlabeled data.

Empirically, patches that cannot be easily distinguished by the local discriminator (e.g., $D_{l}$ is around 0.5) are more likely to correspond to highly transferable features across different diseases, as they capture underlying patterns that remain consistent despite variations in the data distribution. We use the phrase ``transferability'' to capture this property. We define the transferability score of a patch as:
\begin{equation}
\label{transferability_score}
c\left (  \tilde{z}_{ip} \right ) = H\left ( D_{l}\left ( \tilde{z}_{ip}  \right )   \right ) \in \left [ 0, 1 \right ] 
\end{equation}
where $H\left ( \cdot  \right ) $ is the standard entropy function. If the output of the local discriminator $D_{l}$ is around 0.5, then the transferability score is close to 1, indicating that the features in the patch are highly transferable. This enables us to quantify the transferability of patches. Assessing the transferability of patches allows a finer-grained view of the SC, separating an SC into highly transferable and less transferable patches. Features from highly transferable patches will be amplified while features from less transferable patches will be suppressed.

Let $ \mathcal{C}_{i}= \{ {c}_{i1},..., {c}_{ip} \}$ be the transferability scores of patches of SC $i$. The transferability matrix, i.e., the adjacency matrix of the learned transferability-aware graph (TAG), can be formulated as:
\begin{equation}\label{CPgraphGeneration}
A= \frac{1}{B\mathcal{H}}  \sum_{h=1}^{\mathcal{H}}\sum_{i=1}^{B}      \left [  \mathcal{C}_{i}^{T}  \mathcal{C}_{i} \right ] _{\times }
\end{equation}
where $B$ is the batch size, $\mathcal{H}$ is the number of heads, $\left [ \cdot  \right ] _{\times } $ means no gradients back-propagation for the transferability matrix. The adjacency matrix of learned TAG $A$ controls the connection strength of attention between patches.

Since the Transferability Aware Transformer Layer is the final layer in the model, the class token needs to consider the transferability scores and emphasize these high-transferability features from patch tokens. Consequently, the vanilla Self-Attention (SA) in this layer can be upgraded to Transferability Aware Self-Attention (TAS) by incorporating the transferability scores
\begin{equation}\label{SALastLayer}
TAS(q_{cls},K,V) = softmax(\frac{q_{cls}K^{T}}{\sqrt{d} }  )\odot \left [ 1;C\left (  K_{patch}\right )  \right ] V
\end{equation}
where $q_{cls}$ is the query of the class token, $K$ represents the key of all tokens, including the class token and patch tokens, $K_{patch}$ is the key of the patch tokens, $C\left (  K_{patch}\right )$ denotes the transferability scores of the patch tokens, $\odot$ is the dot product, and $\left [ ; \right ] $ is the concatenation operation. The new self-attention encourages the class token to take more information from highly transferable patches while suppressing information from patches with low transferability scores. Accordingly, the multi-head self-attention is therefore renewed as:
\begin{equation}\label{C-MSA}
MTAS(q_{cls},K,V)=Concat(head_{1},...,head_{h} )W^{O}
\end{equation}
where $head_{i}=TAS\left ( q_{cls}W_{i}^{q_{cls}}, KW_{i}^{K} , VW_{i}^{V}   \right ) $, as formulated in Eq.~\ref{SALastLayer}. Taking them together, the operations in the transferability aware transformer layer can be formulated as:
\begin{equation}\label{OpeLL}
\begin{split}
& \hat{z}^{l} =MTAS\left ( LN\left ( \tilde{z}^{l-1}  \right )  \right ) +\tilde{z}^{l-1}\\
& z^{l}=MLP\left ( LN\left ( \hat{z}^{l} \right )  \right ) + \hat{z} ^{l} 
\end{split}
\end{equation}
where $z^{l-1}$ are output from the previous layer. In this way, the Transferability Aware Transformer Layer produces transferability scores for each patch token while also emphasizing fine-grained features that are highly transferable and discriminative for classification. Here $l=L$, $L$ is the total number of transformer layers in ViT architecture.

\subsection{Transferability Graph Guided Transformer Layer}
The adaptively learned TAG can be conveniently integrated to guide self-attention by applying its adjacency matrix via dot product with the attention scores, as depicted in the Transferability Graph Guided Transformer Layer in Fig.~\ref{method}. The TAG guided self-attention focuses on the patches with more transferable features, thus steering the model to capture transferable knowledge. Formally, we can formulate the TAG guided self-attention as:
 \begin{equation}\label{PatchCPSA}
TAG\text{-}SA(Q,K,V,A)=softmax(\frac{QK^T\odot A}{\sqrt{d_k} })V
\end{equation}
where queries, keys, and values of all patches are packed into matrices $Q$, $K$, and $V$, respectively, $A$ is the adjacency matrix of the learned TAG. Accordingly, the TAG guided multi-head attention is then formulated as:
\begin{equation}
\begin{aligned}
MTAG\text{-}SA(Q,K,V,A)=Concat(head_{1}, ...,head_{h})W^{o}\\
\end{aligned}
\end{equation}
where $ head_{i}=TAG\text{-}SA(  QW_{i}^{Q}, KW_{i}^{K} , VW_{i}^{V},A ) $
. The learnable parameter matrices $W_{i}^{Q}$, $W_{i}^{K}$, $W_{i}^{V}$ and $W^{O}$ are the projections. The new knowledge learned from the unseen data is incorporated into the model by integrating the learned matrix with the attention scores.

\subsection{Overall Objective Function}
Since our proposed TAT has a classifier head, a local discriminator, and a global discriminator, there are three terms in the overall objective function. The classification loss is formulated as:
\begin{equation}\label{clcloss}
L_{clc} \left ( x^{s}, y^{s}   \right ) = \frac{1}{n_{s}}\sum_{x_{i}\in D_{s}  }  L_{CE} \left ( G_{c} \left ( G_{f} \left ( x_{i}^{s} \right )  \right ) , y_{i}^{s} \right ) 
\end{equation}
where $G_{c}$ is the classifier head, and $G_{f}$ is the feature extractor of ViT. The global discriminator takes the class token of images and tries to discriminate the class token, i.e., the representation of the entire image, to the labeled support data and unlabeled testing data:
\begin{equation}\label{domainloss}
L_{dis} (x, y^{d})  = -\frac{1}{n} \sum_{x_{i} \in D}L_{ce}\left ( D_{g}\left ( G_{f}\left ( x_{i} \right ) , y_{i}^{d} \right )  \right )  
\end{equation}
where $D{g}$ is the global discriminator, and $y_{i}^{d}$ is the the macro label (i.e., $y_{i}^{d}= 1$ means labeled support data, $y_{i}^{d}= 0$ is the unlabeled testing data). Take classification loss (Eq. \ref{clcloss}), global discrimination loss (Eq. \ref{domainloss}), and local discrimination loss (Eq. \ref{PatchLevelloss}) together, the overall objective function is therefore formulated as:
\begin{equation}
L=L_{clc} \left ( x^{s}, y^{s}   \right ) + \alpha L_{dis}\left ( x, y^{d}  \right ) + \beta L_{pat} (x, y^{d})
\end{equation}
where $\alpha$ and $\beta$ are trainable hyperparameters.

\subsection{Open-set Adaptation}
The label categories in the source dataset include CN, MCI, and AD. However, the label categories in the target dataset consist of CN, MCI, and LBD. This creates an open-set challenge, as the LBD dataset contains a disease category not present in the source dataset.

To address this challenge, we use CN and MCI subjects from the source dataset and include CN, MCI, and LBD subjects from the target dataset during training. Since the labels of subjects from the source dataset are provided during training, the classifier is trained as a binary classifier to distinguish between CN and MCI. To identify the additional LBD disease, we propose a threshold-based entropy approach. For a given subject, the classifier outputs a probability distribution $P=\left [ p_{CN}, p_{MCI}  \right ] $, where $p_{CN}$ and $p_{MCI}$ represent the predicted probabilities for CN and MCI, respectively. We compute the entropy of the prediction as:
\begin{equation}
    H(P)=-p_{CN}\log{p_{CN}} -p_{MCI}\log{p_{MCI}} 
\end{equation}
where $H(P)$ quantifies the uncertainty of the prediction. If the entropy exceeds a predefined threshold $\tau$, the sample is considered to belong to the unknown LBD category:
\begin{equation}
    Predicted \ Label=\left\{\begin{matrix}
 LBD, & if \  H(P) > \tau  \\
 arg \max (P),  & otherwise
\end{matrix}\right.
\end{equation}


\begin{figure}[tb]
\begin{minipage}{0.6\textwidth} 
    \centering
    \captionof{table}{Comparison between TAT and other models on the LBD dataset. Accuracy and standard deviation are reported based on three runs. ViT, TVT, and SSRT only support closed-set adaptation.}
    \begin{tabular}{c|ccc}
        \hline
        Method & CN & MCI & LBD \\ \hline
        ViT & $4.3 \pm 3.5$ & $88.3 \pm 13.5$ & N/A \\
        TVT& $46.4 \pm 5.4$ & $16.7 \pm 13.5$& N/A\\
        SSRT& $47.9 \pm 3.4$& $11.1 \pm 7.9$ & N/A \\ \hline
        TAT &$66.7 \pm 7.4$ & $88.9 \pm 7.9$& $14.5 \pm 4.3$ \\ \hline
    \end{tabular}
    \label{DAResults}
\end{minipage}
\hfill
\begin{minipage}{0.3\textwidth} 
    \centering
\includegraphics[width=\linewidth]{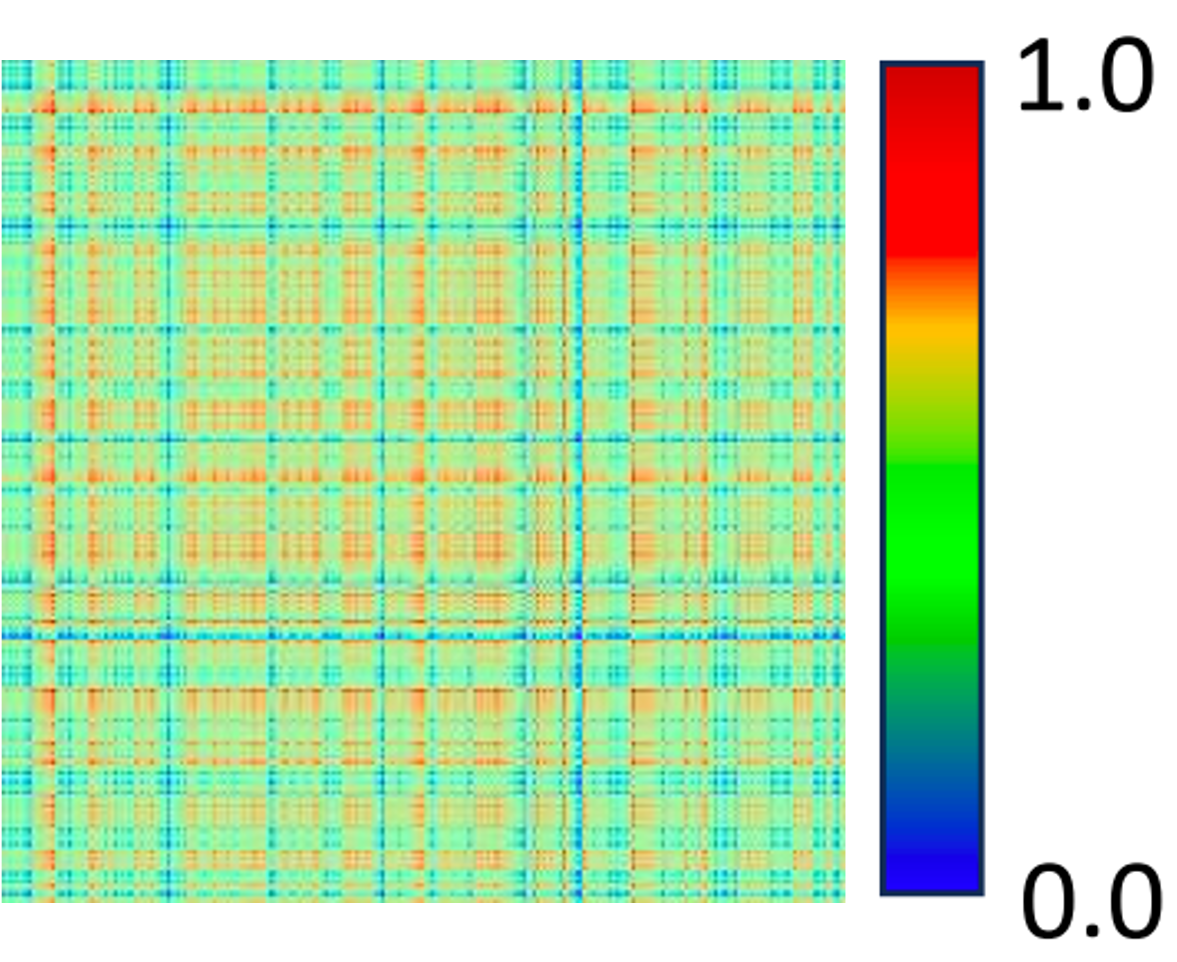}
    \caption{The adjacency matrix of the learned TAG.}
    \label{matricesvis}
\end{minipage}%
\end{figure}

\section{Experiments}


\subsection{Datasets and Training Details}

We use the structural connectivity (SC) constructed from MRI images, and the pre-processing follows the procedures established in existing works~\cite{Zhang2023Representative}. For the AD dataset, we use the public Alzheimer's Disease Neuroimaging Initiative (ADNI) dataset, which is the most widely used Alzheimer's disease dataset~\cite{Yu2023Supervised}. Through SC construction, 282 CN and 149 MCI subjects are available. The LBD dataset was collected using different machines and protocols, resulting in a distinct domain shift from the ADNI dataset. There are 23 CN, 6 MCI, and 77 LBD subjects available, which is significantly fewer than those in ADNI. 

Since the datasets are relatively small in scale, we build our TAT model on ViT-Small (ViT-S) and initialize it with pre-trained weights~\cite{Huang2024Interlude}\cite{Shu2024Real}. We initialize the learning rate at 0 and linearly warm it up to 0.06 over the first 500 training steps, followed by a cosine decay scheduling strategy~\cite{Yu2023Noisynn}\cite{Yu2023Core}. The hyperparameters $\alpha$ and $\beta$ are set to $[1.0, 0.01]$. The entropy threshold is set to $0.8$. We use a mini-batch Stochastic Gradient Descent (SGD) optimizer with a momentum of 0.9. The batch size is set to 16. During training, the model is trained on both the ADNI and LBD datasets. The ADNI samples are provided with labels, whereas the LBD samples are unlabeled. After convergence, we evaluate the prediction results on the LBD dataset against the true labels. This approach enables TAT to adapt to unseen distributions during training.

\begin{table*}[tb]
\centering
\begin{minipage}{0.48\textwidth}
    \centering
    \caption{Ablation study of hyper-parameters of $\alpha$, $\beta$, and $\tau$. Accuracy and standard deviation are based on three runs. }
    \scriptsize
    \begin{tabular}{c|ccc}
        \hline
        ($\alpha$, $\beta$,  $\tau$)     & CN & MCI & LBD \\ \hline
        (1.0,0.1,0.8)        & $57.9 \pm 7.4$ & $88.3 \pm 13.6$ & $21.3 \pm 5.6$   \\
        (0.1,0.01,0.8)  & $60.7 \pm 14.1$   &  $88.3 \pm 13.6$   &   $14.5 \pm 4.3$   \\
        (1.0,0.01,0.7)  & $47.8 \pm 3.6$    &  $77.7 \pm 7.9$    & $54.1 \pm 8.1$   \\ \hline
    \end{tabular}
    \label{HyperAbaltion}
\end{minipage}
\hfill
\begin{minipage}{0.48\textwidth}
    \centering
    \caption{Ablation study of global discriminator (GD) and local discriminator (LD). Accuracy and standard deviation are based on three runs.}
    \scriptsize
    \begin{tabular}{c|ccc}
        \hline
        Method     & CN & MCI & LBD \\ \hline
        TAT   &    $66.7 \pm 7.4$          &  $88.9 \pm 7.9$           &  $14.5 \pm 4.3$  \\
        w/o GD & $50.8 \pm 5.4$   & $66.7 \pm 13.6$    &  $12.9 \pm 2.8$   \\
        w/o LD &  $49.3 \pm 14.7$  &  $88.3 \pm 13.6$   &  $14.2 \pm 4.8$   \\ \hline
    \end{tabular}
    \label{LDGDAblation}
\end{minipage}
\end{table*}

\subsection{Adaptation Results}
The adaptation results are reported in Table~\ref{DAResults}. We compare the proposed TAT with other Transformer-based models. ViT serves as the baseline, while TVT~\cite{Yang2023} and SSRT~\cite{SunSSRT2022} represent the latest advancements in domain adaptation. However, ViT, TVT, and SSRT only support closed-set adaptation, where the label categories are required to be the same between training and inference.
Our proposed TAT not only outperforms these competitors in the MCI and CN categories, but is also able to distinguish the LBD subjects.

We also visualize the learned transferability-aware graph (TAG) to better understand the transferability of the SC patches. The TAG is visualized in the form of the symmetric adjacency matrix, which is shown in Fig.~\ref{matricesvis}. The transferability scores for each SC patch are represented by the intensity of red colors (increasing) and blue colors (decreasing). We observe that different SC patches have different transferability scores. Certain SC patches have higher transferability scores and therefore contain transferable features.


\subsection{Ablation Study}

To evaluate the influence of the hyperparameters and key modules on model performance, we conduct an ablation study. The results are reported in Table~\ref{HyperAbaltion} and~\ref{LDGDAblation}. We observe in Table~\ref{HyperAbaltion} that the model is robust to different combinations of hyperparameters given the limited data samples. A lower entropy threshold leads to higher accuracy in the LBD category, but decreases classification accuracy in others. The results in Table~\ref{LDGDAblation} show that both the global discriminator and local discriminator contribute to the final performance.

\section{Conclusion}

In this work, we propose a novel domain adaptation framework for diagnosing LBD under scarce data and domain shift conditions. TAT addresses two key challenges in LBD analysis: first, it leverages knowledge from the clinically similar AD dataset to enhance model effectiveness; second, it effectively mitigates the distribution shift between datasets collected from different institutions using various machines and protocols. Experimental results demonstrate that TAT outperforms existing domain adaptation methods, achieving robust prediction results for LBD diagnosis.

    

\begin{credits}
\subsubsection{\ackname} This work was supported by the National Institutes of Health (R01AG075582 and RF1NS128534) and funded by the Sir Jules Thorn Charitable Trust (05/JTA), Wellcome Trust Intermediate Clinical Fellowship (WT088441MA), Alzheimer's Research UK Senior Research Fellowship (ARUK-SRF2017B-1), The Lewy body Society (LBS/002/2019) and NIHR Biomedical Research Centres in Sheffield, Cambridge and Newcastle. We also thank Andrew M. Blamire, John-Paul Taylor, Li Su and John T. O’Brien for sharing the data. 

\subsubsection{\discintname}
The authors have no competing interests to declare that are
relevant to the content of this article.
\end{credits}

%
%
%
%

\end{document}